\documentclass[conference]{IEEEtran}
\usepackage{amsmath,amssymb,amsfonts}
\usepackage{graphicx}
\usepackage{textcomp}
\usepackage{xcolor}
\usepackage{booktabs}
\usepackage{tabularx} %
\usepackage{multirow}
\usepackage{subfig}
\usepackage[font=footnotesize]{caption}
\usepackage{xspace}
\usepackage{bm}
\usepackage{balance}
\usepackage[nolist]{acronym}
\usepackage{microtype}

\usepackage[breaklinks,hidelinks]{hyperref}
\usepackage{cleveref} %
\usepackage{fancyhdr}

\usepackage{url}

\def\Snospace~{\S{}}

\renewcommand{\paragraph}[1]{{\vskip 6pt \noindent\textbf{#1.} }}

\definecolor{goeblue}{RGB}{0,51,102}

\usepackage[numbers,square,sort&compress]{natbib}
\def\BibTeX{{\rm B\kern-.05em{\sc i\kern-.025em b}\kern-.08em
    T\kern-.1667em\lower.7ex\hbox{E}\kern-.125emX}}

\newcommand{\dct}{\ensuremath{\mathcal{D}}\xspace}
\newcommand{\idct}{\ensuremath{\mathcal{D}^{-1}}\xspace}
\newcommand{\x}{\ensuremath{\bm{X}}\xspace}
\newcommand{\y}{\ensuremath{\bm{Y}}\xspace}

\newcommand{\G}{\ensuremath{\bm{G}}\xspace}
\newcommand{\R}{\ensuremath{\bm{R}}\xspace}
\newcommand{\F}{\ensuremath{\bm{F}}\xspace}
\newcommand{\M}{\ensuremath{\bm{\tilde{G}}}\xspace}

\newlength{\tempdima}
\newcommand{\rowname}[1]%
{\rotatebox{90}{\makebox[\tempdima][c]{\textit{#1}}}}

\newcommand{\head}[1]{\textbf{#1}} %
\newcommand{\imgnotat}[1]{\footnotesize\textit{#1}}

\begin{document}
	
\fancypagestyle{firststyle}
{
\fancyhf{}
\chead{\small\textit{-------------------------------------- In IEEE 
Deep Learning and Security Workshop (DLS) 2022
--------------------------------------}}
\renewcommand{\headrulewidth}{0pt} }

\title{Misleading Deep-Fake Detection\\with GAN Fingerprints}

\begin{acronym}
	\acro{psnr}[PSNR]{Peak Signal to Noise Ratio}
	\acro{gan}[GAN]{Generative Adversarial Network}
	\acro{dct}[DCT]{discrete cosine transform}
	\acro{lrp}[LRP]{Layer-Wise Relevance Progagation}
\end{acronym}

\author{
	\IEEEauthorblockN{
		Vera Wesselkamp\IEEEauthorrefmark{1},
		Konrad Rieck\IEEEauthorrefmark{1}, 
		Daniel Arp\IEEEauthorrefmark{2} and
		Erwin Quiring\IEEEauthorrefmark{1}
	}
	\IEEEauthorblockA{\IEEEauthorrefmark{1}~\textit{Technische 
	Universit\"at Braunschweig, Germany}}
	\IEEEauthorblockA{\IEEEauthorrefmark{2}~\textit{Technische 
	Universit\"at Berlin, Germany}}
}

\maketitle

\thispagestyle{firststyle}

\begin{abstract}
Generative adversarial networks (GANs) have made remarkable progress in synthesizing realistic-looking images that effectively outsmart even humans. 
Although several detection methods can recognize these deep fakes by checking for image artifacts from the generation process, multiple counterattacks have demonstrated their limitations. These attacks, however, still require certain conditions to hold, such as interacting with the detection method or adjusting the GAN directly.
In this paper, we introduce a novel class of simple counterattacks that overcomes these limitations. In particular, we show that an adversary can remove indicative artifacts, the \emph{GAN fingerprint}, directly from the frequency spectrum of a generated image. 
We explore different realizations of this removal, ranging from filtering high frequencies to more nuanced frequency-peak cleansing.
We evaluate the performance of our attack with different detection methods, GAN architectures, and datasets. Our results show that an adversary can often remove GAN fingerprints and thus evade the detection of generated images.
\end{abstract}

\section{Introduction}
Generative adversarial networks (GANs) are powerful learning models for synthesizing digital media \citep{goodfellow_generative_2014}. They enable  generating images and videos that look astonishingly real. For example, the model StyleGAN can generate portrait photos that are not recognizable as synthetic to the human eye~\cite{karras_analyzing_2020}.
Although GANs have legitimate applications, such as content generation for games and videos~\citep[e.g.,][]{vincentNvidiaHasCreated2018, leeDeepfakeSalvadorDali2019, oakesDeepfakeVoiceTech2019}, their ability to create forged images---so called \emph{deep fakes}---resembles a prime tool for misuse, for example, as part of propaganda and disinformation campaigns~\cite{osullivanHighSchoolStudent2020, chiuFacebookWouldnDelete2019, vahiaDeepfakeBotsCreate2020}. 

Prior work has successfully established different methods for detecting deep-fake images using unique artifacts that GANs leave in the data~\cite[e.g.,][]{yu_attributing_2019, frank_leveraging_2020, joslin_attributing_2020, marra_gans_2019, zhang_detecting_2019, wang_cnn-generated_2020}.
In particular, the frequency domain of images has proven to be useful for this task, allowing an almost perfect detection~\citep{frank_leveraging_2020}. As a result of this performance, different counterattacks have been developed that allow evading the detection of generated images~\cite{carlini_evading_2020, cozzolino_spoc_2019, huang_fakepolisher_2020}. However, from the adversary's perspective, these attacks still require certain conditions to hold, such as interaction with the detection method or direct adaptation of the GAN model, which limits their practicality.

\begin{figure}[b]
	\centering
	\includegraphics{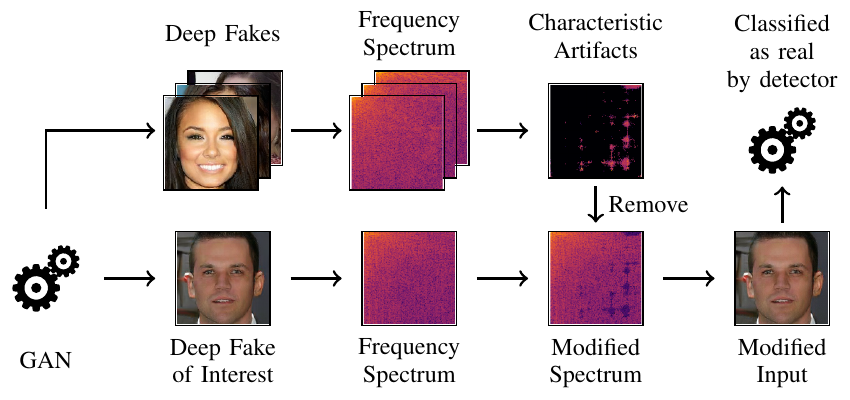}
	\vspace{-0.150cm}
	\caption{Illustration of our counterattacks. The adversary calculates the characteristic GAN artifacts in the frequency spectrum and removes this fingerprint to avoid detection.}
	\label{fig:overview-compact}
\end{figure}

In this paper, we introduce a novel class of simple counterattacks that overcomes these limitations. 
These attacks build on the concept of a \emph{GAN fingerprint}, a consistent frequency pattern that characterizes the generation process similar to a camera fingerprint in digital forensics. By identifying and removing this fingerprint from generated images, our attack obstructs frequency-based detection approaches. The fingerprint removal requires no adaption of the GAN model and is agnostic to the detection method. \autoref{fig:overview-compact} illustrates this concept: The adversary first generates multiple images, estimates the resulting GAN fingerprint (upper~row), and finally removes it from a target image (lower row).

The removal of a GAN fingerprint, however, is not a trivial task, as generation artifacts manifest in different frequency bands and patterns. As a consequence, we develop four variants of our attack, gradually increasing their sophistication. We start by simply removing high frequencies from images. This variant is surprisingly effective if the GAN fingerprint is located in high-frequency bands, yet it also affects image details. As a remedy, the second variant targets the fingerprint more precisely by removing the differences between the mean frequency spectra of fake and natural images. The third variant refines this approach and only removes peaks from the frequency differences. Finally, the last variant uses a regression model to estimate discriminative patterns in the frequency spectra.

We empirically evaluate the performance of these four attack variants with different detection methods, GAN architectures, and datasets. In particular, we employ the detection method by \citet{joslin_attributing_2020} and two learning-based classifiers by \citet{frank_leveraging_2020}. Our evaluation shows that the removal of GAN fingerprints misleads all detection methods. While the mean-spectrum attack is highly effective against \citeauthor{joslin_attributing_2020}, the removal of high frequencies or frequency peaks evades \citeauthor{frank_leveraging_2020} in most cases.
Contrary to our expectations, these simple attack variants are more effective than our learning-based regression attack. All in all, our findings demonstrate that adversaries can evade detection methods with relatively simple means and there is a need for more robust concepts.

\paragraph{Contributions}
In summary, our contributions are as follows:

\begin{itemize} \vspace{1ex}
    \setlength{\itemsep}{4pt}
    \item \emph{GAN fingerprints against deep-fake detection.} 
    We show that removing the characteristic artifacts of GAN images in the frequency spectrum is a simple yet effective counterattack against deep-fake detection methods. 
    \item \emph{Manipulation strategies.}
    We present four methods for modifying the frequency spectrum. They range from removing \mbox{high frequencies} to more nuanced artifact removals.
    \item \emph{Comprehensive evaluation.}
    We empirically evaluate our attacks on three detection methods, four GAN architectures, and two datasets. The detection rate from each GAN can be considerably reduced by one of our attacks.
\end{itemize}

We make the source code and dataset information available under: {\small \textcolor{goeblue}{\url{https://github.com/vwesselkamp/deepfake-fingerprint-attacks}}}.

\section{Deep Fake Detection}
\label{sec:deepfakedetection}
Approaches for detecting deep-fake images can be broadly divided into two groups: %
The first group checks the consistency of an image. 
For instance, inconsistent physical traits can be leveraged~\cite{verdoliva_media_2020}, 
such as the pose of the head or facial symmetry of eyes and earrings. 
Likewise, the color saturation or other disparities in the
color components of images can also uncover a deep fake~\cite{verdoliva_media_2020}. 
The second group relies on (invisible) image artifacts that the generation process introduces~\cite{joslin_attributing_2020, frank_leveraging_2020}. 
Their advantage is that artifacts can be automatically derived for each GAN. This allows for a rather generic identification. Recent work also suggests that artifacts may even transfer between different GANs~\cite{wang_cnn-generated_2020}. In this paper, we focus on these artifact-based approaches.

\subsection{GAN Artifacts and Fingerprints}

\begin{figure}[]
	\settoheight{\tempdima}{\includegraphics[width=.22\linewidth]{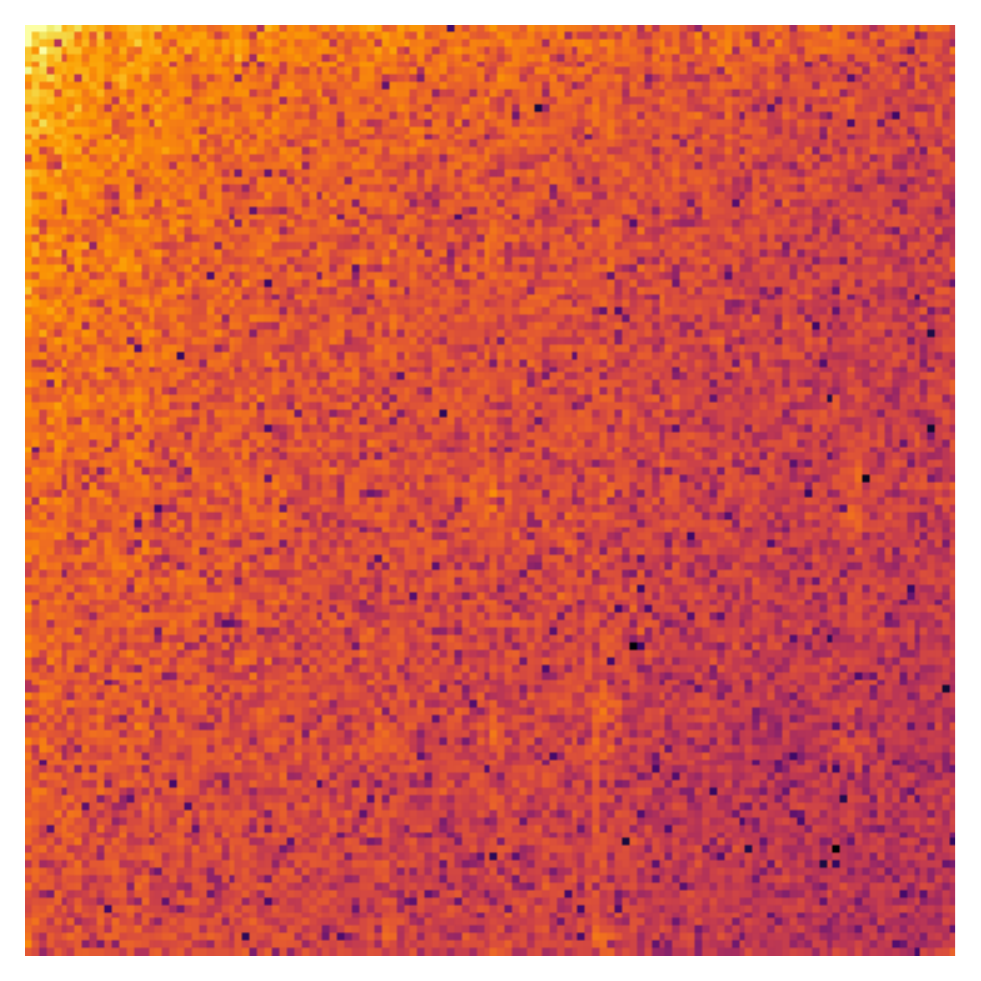}}%
	\centering
	\begin{tabular}{c@{ }c@{ }c@{ }c@{ }c@{ }c@{}}
		\imgnotat{Natural} &\imgnotat{ProGAN} & \imgnotat{SNGAN} & \imgnotat{CramerGAN} & 
		\imgnotat{MMDGAN}\\
		\includegraphics[width=.18\linewidth]{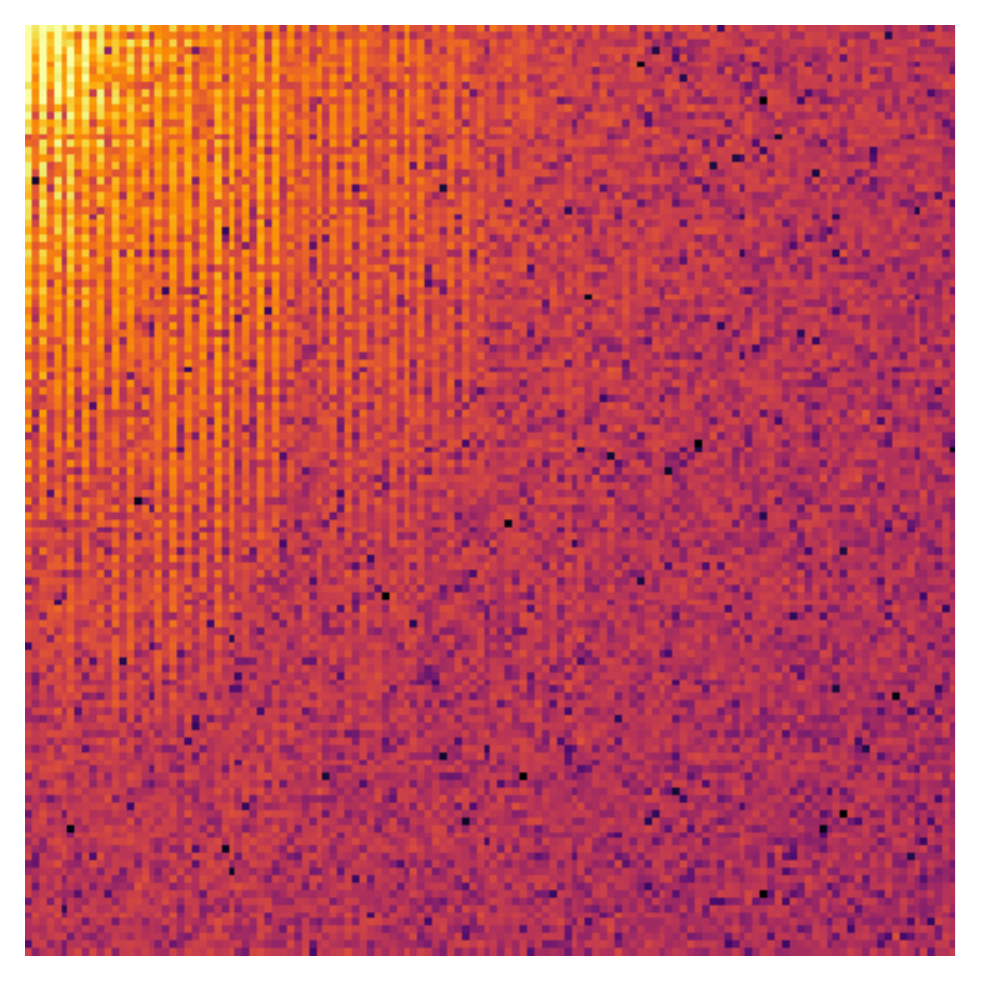}&
		\includegraphics[width=.18\linewidth]{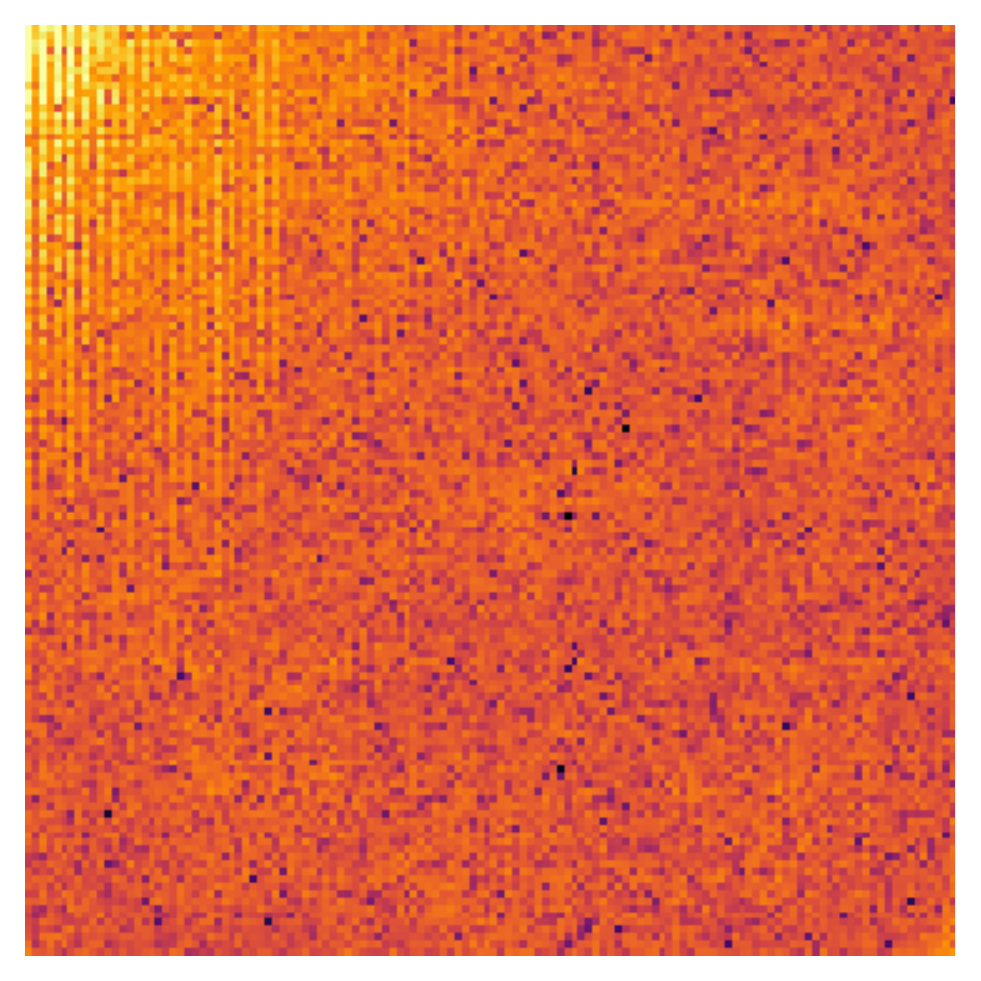}&
		\includegraphics[width=.18\linewidth]{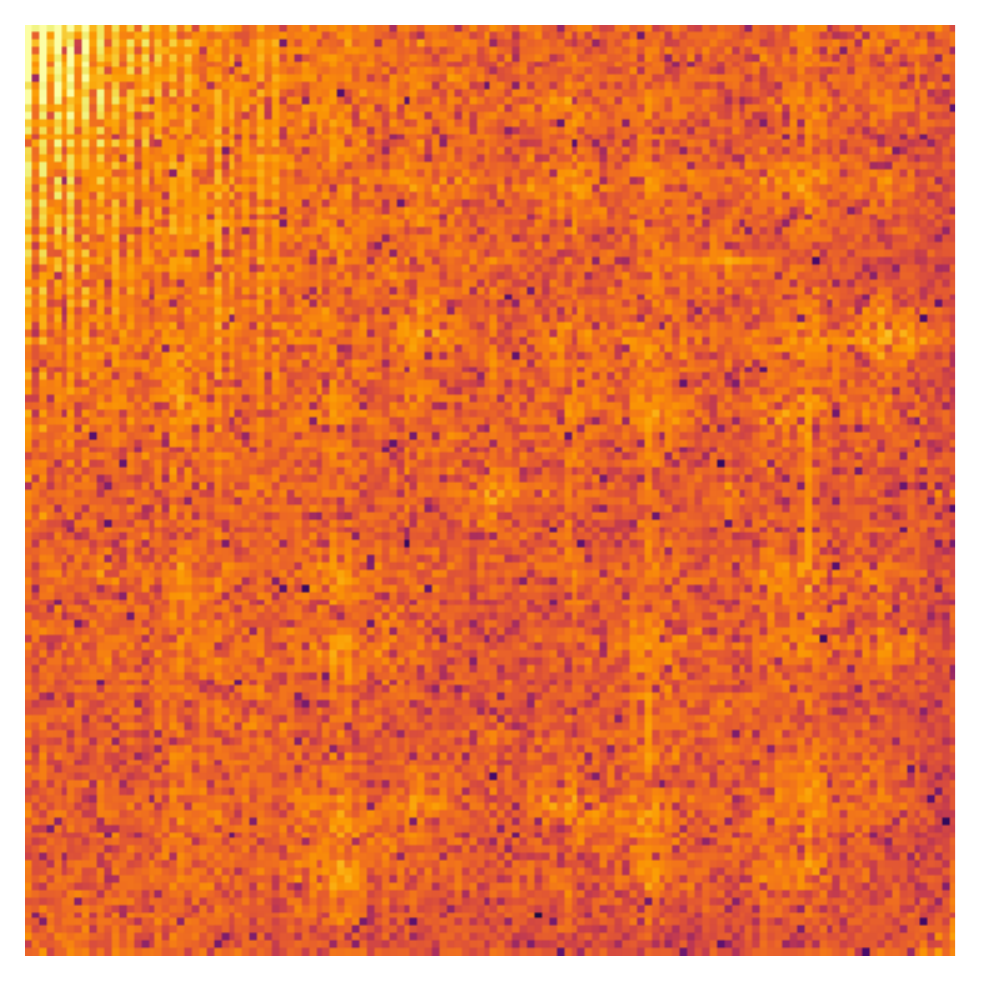}&
		\includegraphics[width=.18\linewidth]{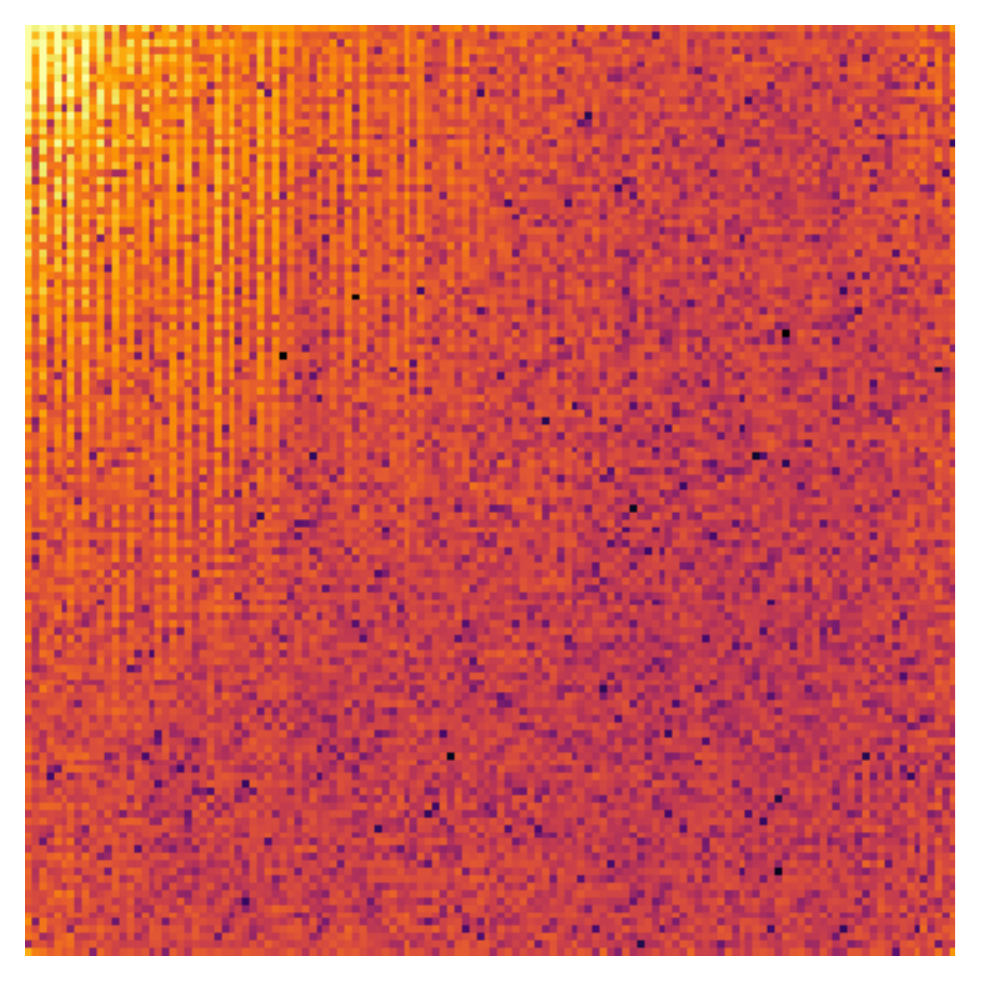}&
		\includegraphics[width=.18\linewidth]{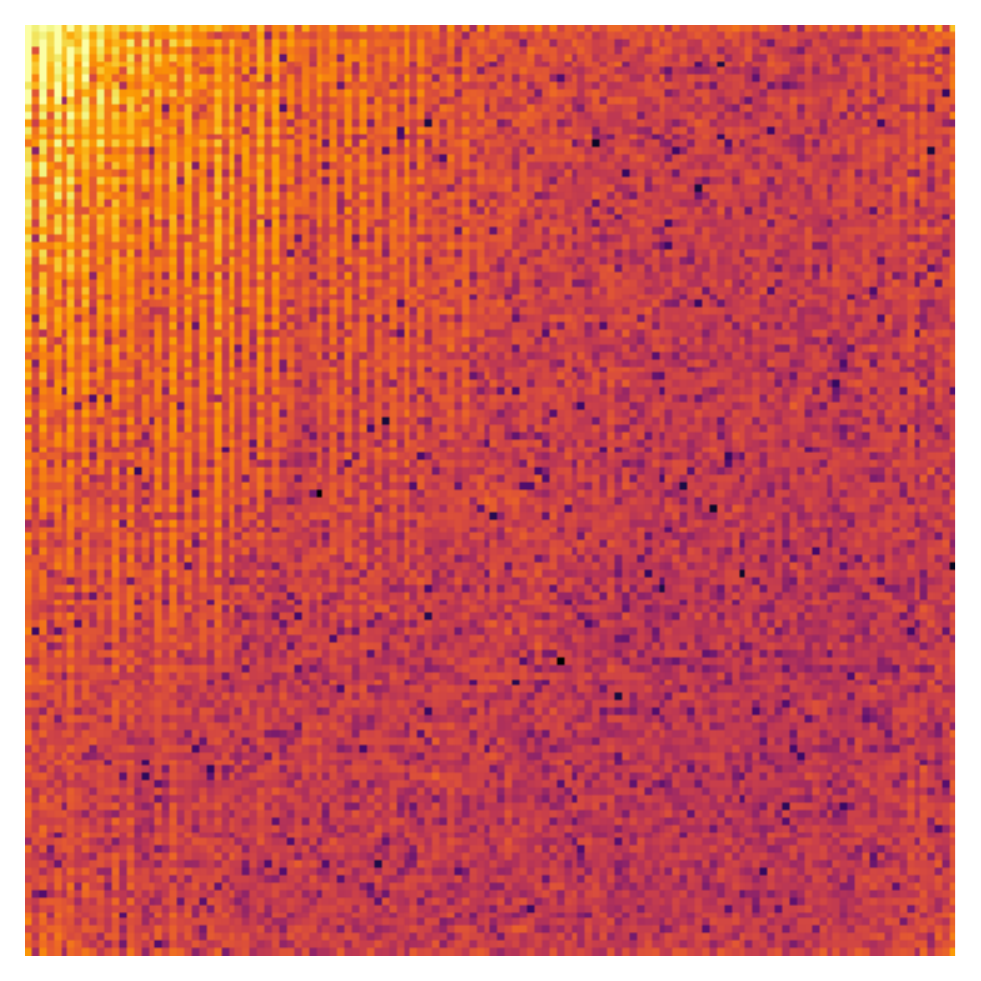}\\[-1ex]
	\end{tabular}	
	\caption{Mean DCT spectra from real CelebA images and from four GANs on the CelebA dataset. We average the DCT spectrum of 5000 images, log-scale the mean, and cut it to [-10,10], respectively.} %
	\label{fig:example-spectrum}
\end{figure}

To provide a first intuition, \autoref{fig:example-spectrum} shows the averaged discrete cosine transform (DCT) spectrum from natural and GAN-generated images, respectively. Two aspects are noticeable: (a) GAN images lead to visible, characteristic artifacts in the frequency spectrum, and (b) these artifacts vary between the different GAN models. For instance, SNGAN induces a grid-like pattern while ProGAN leads to higher values across all frequencies. This simple example underlines that there are clear patterns that differentiate real from GAN images.

The existence of GAN-specific artifacts has been attributed to the up-sampling operations when increasing image resolution~\cite{frank_leveraging_2020, odena_deconvolution_2016, zhang_detecting_2019}.
Initially, GANs for image generation~\cite{miyato_spectral_2018, binkowski_demystifying_2018, bellemare_cramer_2017} 
used transposed convolution in their up-sampling, which leads to checkerboard artifacts in the spatial domain of images. This occurs when the kernel size is not divisible by the stride by which the kernel moves 
over the pixels of the low-resolution image. The artifacts 
created in one layer thus accumulate over several layers and result 
in patterns in the final image~\cite{odena_deconvolution_2016}.
Hence, recently proposed GANs, such as  ProGAN~\cite{karras_progressive_2018}, switched to interpolation 
followed by convolution.  
While using an interpolation during up-sampling does not produce strong artifacts in the  spatial domain anymore, \citeauthor{frank_leveraging_2020} show that  different kinds of interpolation still lead to detectable patterns  in the frequency domain~\cite{frank_leveraging_2020}.

These frequency artifacts can be denoted as a \emph{GAN fingerprint}~\cite{marra_gans_2019, joslin_attributing_2020}, as they are consistently present in images from the same GAN model, but differ between images from different GAN models, similar to a camera fingerprint in digital forensics. %
This view motivates our counterattacks in Section~\ref{sec:counterattacks} that aim at removing or suppressing a GAN~fingerprint to bypass a deep-fake detection.

\subsection{Detection Methods}
\label{sec:detection-methods}
Artifact-based approaches can be further divided into two subgroups: they operate either in the spatial domain~\cite{marra_gans_2019, yu_attributing_2019} or in the frequency domain~\cite{frank_leveraging_2020,joslin_attributing_2020, zhang_detecting_2019, qianThinkingFrequencyFace2020, durall_unmasking_2020, guarneraPreliminaryForensicsAnalysis2020}. 
A recent comparison by \citet{frank_leveraging_2020} demonstrates multiple advantages of frequency-based approaches, such as a higher accuracy and robustness against image perturbations. For our evaluation, we thus focus on frequency-based approaches and implement the following two detection methods. 

First, we consider the fingerprint method by \citet{joslin_attributing_2020}. It basically computes a fingerprint by averaging the FFT frequency spectrum of a set of GAN images. The detection is based on computing the cosine similarity between the fingerprint and the FFT spectrum of the image under investigation. 
Second, we examine the learning-based method by \citet{frank_leveraging_2020}. We consider two models: a Ridge regression and a CNN. Both are trained on the DCT frequency spectrum from natural and generated images. The CNN differentiates five classes (natural images and 4 GAN models), while the regression is a binary classifier that is trained for each GAN individually (see \autoref{sec:evaluation}).
We choose the regression, since the weights of a regression model have been demonstrated to correspond to periodic patterns in the frequency spectrum. This motivates our fingerprint-based counterattacks that suppress these frequency patterns. The CNN classifier provides the highest detection rate in prior work and thus allows us to test our counterattacks against the current state of the art~\cite{frank_leveraging_2020}.

\section{Counterattacks}
\label{sec:counterattacks}
We proceed to introduce our novel class of counterattacks. These attacks build on the concept of GAN fingerprints: If a characteristic pattern is present in all generated images, an attacker can try to remove this pattern to evade detection. Such an attack is rather simple to realize. The adversary only has to modify the generated image---adjusting the GAN model is not necessary. Also, the adversary neither requires detailed knowledge of the detection method nor needs to interact with it. As a result, our counterattacks are easy to employ in practice using existing GAN models for generation.

There is, however, a crux: Our evaluation shows that there is \emph{no universal} fingerprint for a GAN that can be simply removed to fool all detection approaches. Instead, each detection method makes use of a different subset of artifacts that affect the detection of fingerprints. 
Therefore, we derive four different variants of our counterattack with increasing  complexity. We start by disturbing the fingerprint through the removal of high frequencies (\autoref{sec:fingerprint-disturbance}) and continue to gradually focus this removal on specific frequency patterns (\autoref{sec:fingerprint-removal}).

\paragraph{Notation}
Matrices and vectors are written in boldface font. If not stated 
otherwise, operations on matrices are point-wise. 
We denote the DCT transformation of a spatial signal~\x by $\dct(\x) 
= \y$, the inverse DCT by $\idct(\y) = \x$. Furthermore, \G denotes 
GAN-generated images, \R real images, \F fingerprints, and \M 
manipulated GAN images.

\paragraph{Threat Model}
We assume a black-box scenario for a counterattack.
The adversary has access to a GAN model and uses it to generate deep-fake images. A defender aims at identifying these images using a detection method.
The adversary has no inner knowledge of this detection method and cannot interact with the method.
Finally, we assume that adapting the GAN model is costly for the adversary. As a result, she focuses on attacks that manipulate the generated images only.

\subsection{Untargeted Fingerprint Removal}
\label{sec:fingerprint-disturbance}
Motivated by prior work that establishes the importance of high 
frequencies for the detection of deep fakes~\cite{odena_deconvolution_2016, 
yu_attributing_2019, joslin_attributing_2020, durall_unmasking_2020}, 
our first attack variant simply removes the high frequency 
spectrum. In particular, we apply an ideal low-pass filter and set bars of width $s$ of DCT coefficients along the lower and right edges of the spectrum to zero. 
Figure~\ref{fig:all-fingerprints} exemplifies this attack, which we refer to as \emph{frequency-bars attack}. %

The attack filters high frequencies from the images. These correspond to details that are less visible for humans and are typically first removed by image compression methods, such as JPEG compression. The intended effect of our attack is similar to \textit{blurring}, which is a typical baseline attack for evading detection in the literature~\cite{yu_attributing_2019, frank_leveraging_2020, joslin_attributing_2020}. Yet, the size of the bars~$s$ in our attack allows a finer control over the removed information as we demonstrate in \autoref{sec:evaluation}. 

Although this attack is straightforward to realize and does not require the fingerprint itself, it induces some drawbacks. 
The attack affects both the fingerprint and the actual image. 
Moreover, it does not entirely clean a deep fake from artifacts if parts of the fingerprint are located in lower frequencies.
As a remedy, we develop more \emph{target-oriented attacks} in the next section that aim at the actual fingerprint.

\subsection{Targeted Fingerprint Removal}
\label{sec:fingerprint-removal}
\begin{figure}[]
	\subfloat[Frequency bars]
    {\includegraphics[width=.235\linewidth]{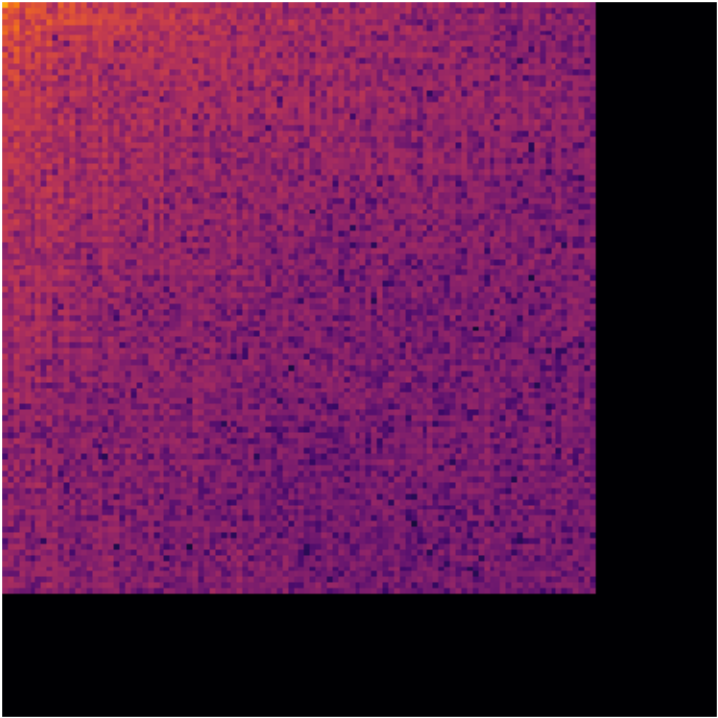}}\hfill
	\subfloat[Mean spectrum]
	{\includegraphics[width=.235\linewidth]{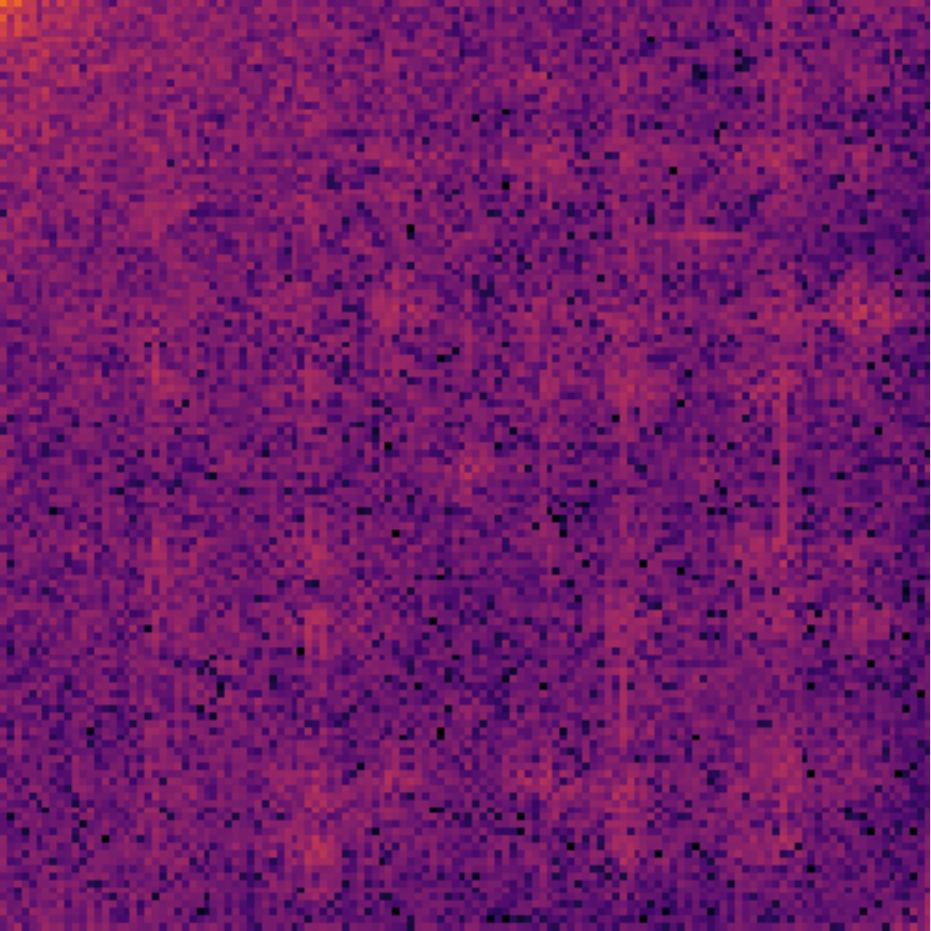}}\hfill
	\subfloat[Peaks]
	{\includegraphics[width=.235\linewidth]{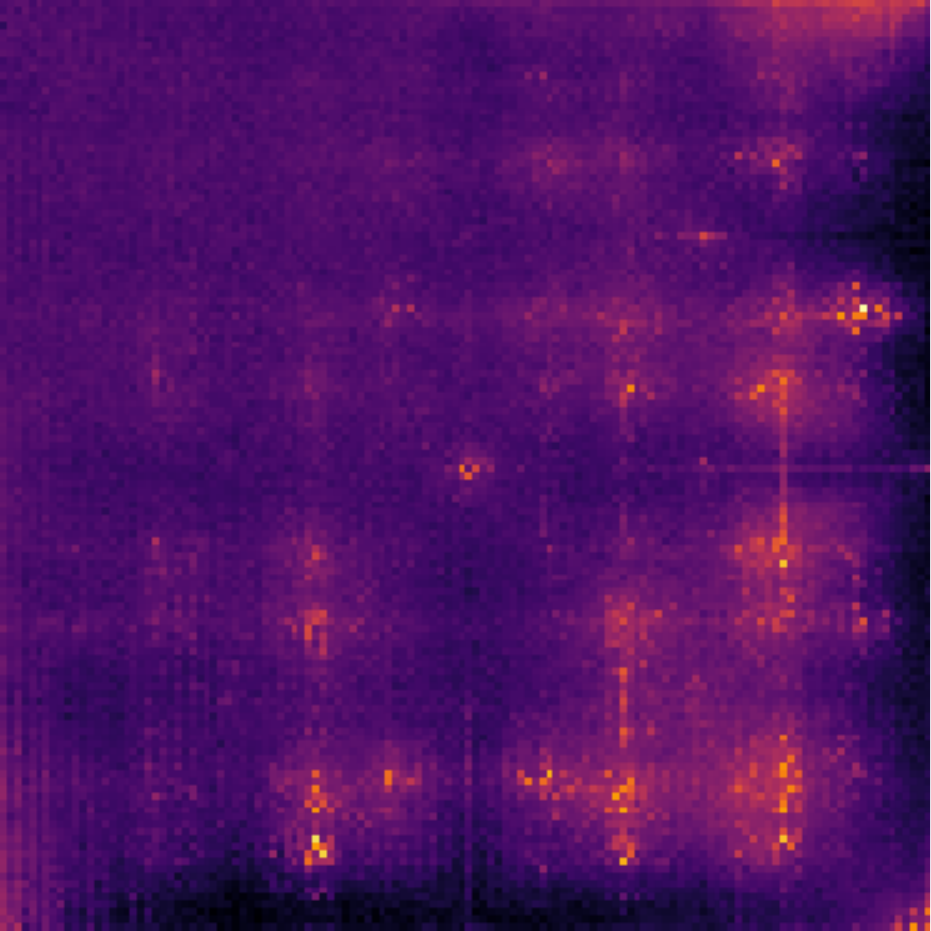}}\hfill
	\subfloat[Regression]
	{\includegraphics[width=.235\linewidth]{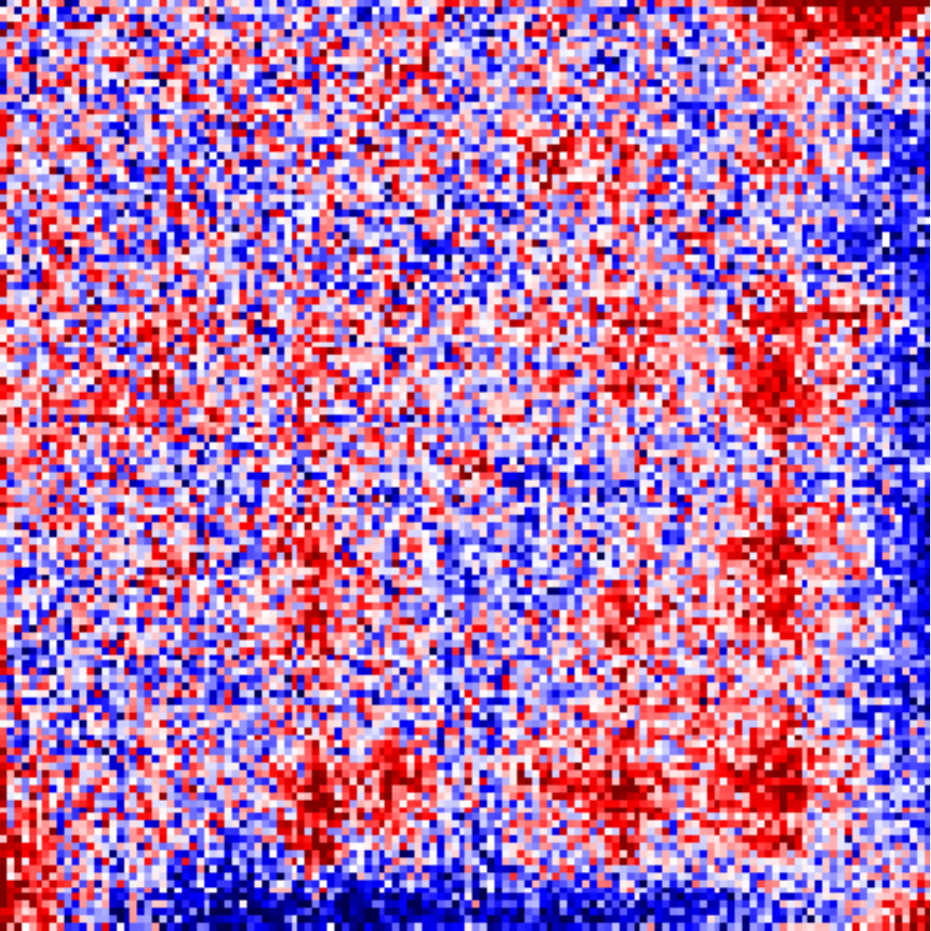}}\hfill
	\caption{Counterattacks for CelebA SNGAN. Plot (a) shows the removal of high-frequency bands; Plot (b)--(d) show the fingerprints that are suppressed. Note that plot (c) shows the fingerprint before applying the threshold.}
	\label{fig:all-fingerprints}
\end{figure}

We present three attack variants that extract the frequency fingerprint for a GAN model and then suppress it in generated images of this GAN.
\autoref{fig:all-fingerprints} exemplifies the fingerprints of the presented attacks for the CelebA SNGAN model. 

\paragraph{Mean-Spectrum Attack} For this attack, we calculate the difference in the respective mean spectra of natural images and GAN-generated images to determine a fingerprint. 
\begin{equation}
    \F_m = \frac{1}{n}\sum_{i=0}^{n} \dct(\G_i) - 
    \frac{1}{n}\sum_{i=0}^{n} \dct(\R_i)
\end{equation}
As counterattack, we simply subtract the mean fingerprint $\F_m$ from 
a GAN-generated image with strength $s$:
\begin{equation}
	\label{eqn:abs-removal}
    \M_i = \idct(\dct(\G_i) - s \cdot \F_m)
\end{equation} 

\paragraph{Frequency-Peaks Attack}
Prior work shows that GAN artifacts are often 
visible in the frequency domain of images as periodic peaks~\cite{frank_leveraging_2020}. We 
attempt to target these peaks directly by only manipulating the 
frequency coefficients above a certain threshold.
To this end, we again compute the mean spectrum, but now on log-scaled values. 
As the DCT of an image leads to larger coefficients for low frequencies, log-scaling reduces the emphasis on the low frequencies. %
We finally execute our manipulations on the non-log-scaled DCT-spectra of GAN-generated images, so that we need to exponentiate the difference. Our \textit{peak} fingerprint $\F_p$ becomes:
\begin{equation}
    \F_p = \exp \left ( \frac{1}{n}\sum_{i=0}^{n} \log(\dct(\G_i)) - 
    \frac{1}{n}\sum_{i=0}^{n} \log(\dct(\R_i)) \right )
\end{equation}
In this way, frequency patterns become more pronounced in the fingerprint (see~\autoref{fig:all-fingerprints}).
We target only the most dominant parts of the pattern: We scale $\F_p$ to $[0,1]$, apply binary thresholding which keeps values larger than a threshold~$t$ and sets smaller values to $0$, 
then intensify the kept values with a strength parameter $s$, and finally clip values to $[0,1]$ again. 
The latter avoids switching signs during fingerprint removal. %
The attack is then given as:
\begin{align}
    \M_i &= \idct (\dct(\G_i)  ( 1- \tilde{\F_p} ) ) \\
  \mathrm{with} \; \tilde{\F_p} &= \mathrm{clip}(s \cdot \mathrm{threshold}(\mathrm{scale}(\F_p), t)) \nonumber 
    \label{eqn:prop-removal}
\end{align}
Note that all operations are element-wise. Different from 
Equation~\ref{eqn:abs-removal}, the multiplication reduces the 
coefficients of the DCT spectrum proportionally to the strength of 
the fingerprint.

\paragraph{Regression-Weights Attack}
For the fourth attack variant, 
we estimate a fingerprint from weights learned by a regression model. We choose a Lasso regression here, since it pushes the weights of features with little influence on the output towards zero, thus effectively extracting the most relevant features for classification. 
Moreover, the weights have a direct correspondence to the frequency coefficients, so that a counterattack can directly change the coefficients anti-proportionally to the respective weights. 
If $\F_r$ denotes the regression weights, the counterattack is defined as:  
\begin{equation}
    \M_i = \idct(\dct(\G_i) ( 1- \tilde{\F_r} ) )
    \label{eqn:proportional-removal}
\end{equation}
where $\tilde{\F_r}$ is clipped, that is,  $\tilde{\F_r} = \mathrm{clip}(s * \F_r)$ with $\mathrm{clip}$ reducing the range to $[-1, 1]$.

\section{Evaluation}
\label{sec:evaluation}
We proceed to empirically evaluate our counterattacks against deep-fake detection methods. First, we show that the detection rate of deep fakes from each GAN model can be considerably reduced by one of the attack variants (\autoref{subsec:eval-attack-success}) while having only a minor visible impact on the image (\autoref{subsec:eval-image-quality}). Second, we demonstrate that our counterattacks achieve a higher success rate than previously used perturbation-based attacks (\autoref{subsec:evaluation-image-perturbations}).

\subsection{Experimental Settings} 

\paragraph{Dataset and GAN Models} 
We adopt the experimental setup from prior work~\cite{yu_attributing_2019, frank_leveraging_2020}:
we evaluate four GAN architectures (ProGAN, SNGAN, MMDGAN, CramerGAN) where each is trained on two datasets of natural images (CelebA~\cite{liu_large-scale_2015} and LSUN bedrooms~\cite{yu15lsun}), respectively.
In total, this setup leads to 8 different combinations of architecture and dataset. 
The images have a size of $128 \times 128 \times 3$ pixels. Further information about the dataset can be found in our github paper repository. 
\paragraph{Deepfake Detectors} 
As described in \autoref{sec:deepfakedetection}, we consider multiple detection methods. \autoref{tab:evaluation-setup-detection} summarizes the setup. 
Note that we obtain one detector for each dataset in the multi-class setting, while a binary classifier requires the creation of a single detector for each combination of architecture and dataset.

\begin{table}[ht]
	\centering 
	\begin{tabularx}{0.975\columnwidth}{llX}
		\toprule \head{Detection} & \head{Type} & \head{Domain} \\ \midrule 
		\citet{joslin_attributing_2020} & Binary & Frequency (Fourier) \\
		\citet{frank_leveraging_2020} CNN & Multi-class & Frequency (DCT)          \\
		\citet{frank_leveraging_2020} Regression & Binary  & Frequency (DCT)            \\ \bottomrule 
	\end{tabularx}
	\caption{Detection setup. Multi-class has five classes $\lbrace$ProGAN, SNGAN, CramerGAN, MMDGAN, Natural$\rbrace$.}
	\label{tab:evaluation-setup-detection} 
\end{table}

To assess the efficacy of our counterattacks, we first compute the accuracy of the detection methods for unmodified deep-fake images. \autoref{tab:attack-success} presents the accuracy for each setup. While the approach by \citet{frank_leveraging_2020} exhibits an almost perfect detection rate, the performance of \citet{joslin_attributing_2020} varies significantly for different GAN architectures, yielding the best detection rate for SNGAN.

\paragraph{Calibrating Fingerprints} 
\label{sec:counter-attacks-impl} 
We extract the fingerprints for our attacks on a separate hold-out dataset. 
The threshold~$t$ for the frequency-peaks attack is determined for each GAN model on this set through a simple grid search.
For the regression-weights attack, we retrieve the weights for the fingerprint by training a Lasso regression on the hold-out dataset.

\paragraph{Evaluation Measures} We evaluate the performance of counterattacks in terms of \emph{attack success rate} and \emph{image quality}. In particular, we measure the attack performance as the fraction of generated images classified as natural. Note that we aim at a targeted attack in the multi-class setting: an attack is only counted as successful if the detection method misclassifies an image as natural rather than just assigning the wrong GAN class.
Furthermore, we measure the visual quality in terms of the \ac{psnr}, which is a commonly used metric in image processing~\cite{SenMem13}. After visual inspection, we consider a \ac{psnr} value of 30dB as an acceptable lower bound for the image quality.

\subsection{Attack Success Rate} 
\label{subsec:eval-attack-success}

In the first experiment, we investigate whether the presented counterattacks allow modifying a deep fake such that it is misclassified as a natural image. To this end, we apply the counterattacks on 1,000~images from each GAN model against the three detection methods. Each attack is calibrated using the strength $s$ so that the average \ac{psnr} of the 1,000 manipulated images is 30dB.

\begin{table*}[] 
	\centering 
\resizebox{\textwidth}{!}{\begin{tabular}{@{}lllr|rrrr|rrrr@{}}
\toprule
       &            &           &                         & \multicolumn{4}{c|}{Our counterattacks (success rate)}                                    & \multicolumn{4}{c}{Baseline perturbations (success rate)}            \\
Dataset& Detection  & GAN Model & Accuracy                & Frequency bars  & Mean spectrum     & Peak Extraction & Regression      & Cropping         & Noise   & Blurring       & JPEG    \\ \midrule
LSUN   & Joslin     & ProGAN    & 56.7\%                  & 69.20\%         & \textbf{96.4\%}   & 71.60\%         & 65.80\%         & 75.70\%          & 74.20\% & 76.40\%        & 75.50\% \\
       &            & SNGAN     & 97.8\%                  & 13.40\%         & 73.5\%            & 4.70\%          & 4.40\%          & \textbf{95.60\%} & 10.70\% & 20.30\%        & 17.10\% \\
       &            & CramerGAN & 55.5\%                  & 50.80\%         & \textbf{91.6\%}   & 48.10\%         & 47.80\%         & 55.20\%          & 52.90\% & 56.20\%        & 56.80\% \\
       &            & MMDGAN    & 57.4\%                  & 47.00\%         & \textbf{82.6\%}   & 42.90\%         & 41.90\%         & 54.00\%          & 47.00\% & 49.70\%        & 50.50\% \\ \midrule
       & CNN        & ProGAN    & \multirow{4}{*}{99.0\%} & 89.6\%          & 0\%               & \textbf{92\%}   & 0.1\%           & 12.7\%           & 0\%     & 54.2\%         & 25.2\%  \\
       &            & SNGAN     &                         & \textbf{91.8\%} & 0\%               & 1.4\%           & 0\%             & 7.3\%            & 0\%     & 56.7\%         & 10.1\%  \\
       &            & CramerGAN &                         & \textbf{91.1\%} & 0\%               & 0\%             & 0\%             & 0.3\%            & 0\%     & 62.9\%         & 8.7\%   \\
       &            & MMDGAN    &                         & \textbf{90.8\%} & 0\%               & 0\%             & 0\%             & 0.2\%            & 0\%     & 56.1\%         & 13.2\%  \\ \midrule
       & Regression & ProGAN    & 91.8\%                  & \textbf{100\%}  & 10.4\%            & \textbf{100\%}  & 32.9\%          & 5.1\%            & 82.6\%  & \textbf{100\%} & 61.5\%  \\
       &            & SNGAN     & 98.9\%                  & \textbf{100\%}  & 0\%               & \textbf{100\%}  & 1.7\%           & 24.1\%           & 25.8\%  & 95.3\%         & 13.2\%  \\
       &            & CramerGAN & 99.1\%                  & \textbf{100\%}  & 0\%               & 2.9\%           & 7.9\%           & 35.5\%           & 49.5\%  & 99.5\%         & 80.8\%  \\
       &            & MMDGAN    & 99.3\%                  & \textbf{100\%}  & 0.4\%             & 1.2\%           & 57.6\%          & 71.5\%           & 47.7\%  & 99.9\%         & 91\%    \\ \midrule
CelebA & Joslin     & ProGAN    & 79.2\%                  & 83.40\%         & \textbf{100.00\%} & 29.50\%         & 28.40\%         & 84.50\%          & 43.80\% & 72.20\%        & 69.00\% \\
       &            & SNGAN     & 95.9\%                  & 85.20\%         & \textbf{99.60\%}  & 13.20\%         & 6.40\%          & 96.10\%          & 20.40\% & 68.00\%        & 66.40\% \\
       &            & CramerGAN & 61.3\%                  & 73.80\%         & \textbf{95.40\%}  & 53.30\%         & 53.00\%         & 80.80\%          & 61.40\% & 71.70\%        & 69.20\% \\
       &            & MMDGAN    & 57.8\%                  & 70.30\%         & \textbf{92.30\%}  & 69.10\%         & 69.10\%         & 85.80\%          & 76.90\% & 78.50\%        & 79.30\% \\ \midrule
       & CNN        & ProGAN    & \multirow{4}{*}{99.3\%} & 98.2\%          & 0\%               & \textbf{99.9\%} & 17.9\%          & 8.4\%            & 0\%     & \textbf{100\%} & 2.7\%   \\
       &            & SNGAN     &                         & \textbf{100\%}  & 0\%               & \textbf{100\%}  & 1.4\%           & 3.8\%            & 0\%     & \textbf{100\%} & 1.5\%   \\
       &            & CramerGAN &                         & 93.1\%          & 0\%               & 0.8\%           & 0\%             & 10.5\%           & 0\%     & \textbf{100\%} & 2.4\%   \\
       &            & MMDGAN    &                         & 99.5\%          & 0\%               & 0\%             & 0\%             & 25.9\%           & 0.1\%   & \textbf{100\%} & 3.2\%   \\ \midrule
       & Regression & ProGAN    & 93.3\%                  & 20.8\%          & 0.2\%             & \textbf{100\%}  & 73.3\%          & 13.8\%           & 76.2\%  & 85.1\%         & 56.7\%  \\
       &            & SNGAN     & 96.7\%                  & 64.7\%          & 0\%               & \textbf{100\%}  & 0.7\%           & 0.6\%            & 60.5\%  & 72.9\%         & 22.4\%  \\
       &            & CramerGAN & 97.4\%                  & \textbf{100\%}  & 0.8\%             & 72.1\%          & 99.9\%          & 53.4\%           & 36.7\%  & 84.2\%         & 47.8\%  \\
       &            & MMDGAN    & 97.3\%                  & 97.7\%          & 2.2\%             & 99.1\%          & \textbf{99.4\%} & 39.1\%           & 38.1\%  & 83.4\%         & 87.1\%  \\ \bottomrule
\end{tabular}}
		\caption{
		The accuracy of deep-fake detection and the success rate of our counterattacks \& baseline perturbations for evading the detection---per dataset, detection method, and GAN model.
		The detection accuracy is computed on 1,000 natural \& 1,000 generated images with a binary classifier, and 1,000 natural \& 4,000 generated images (1,000 of each GAN model) in a multi-class case. 
		In terms of image quality, the attacks are calibrated to a \ac{psnr} value of~30~dB.}
		\label{tab:attack-success} 
\end{table*}

\paragraph{Results}
\autoref{tab:attack-success} shows the performance of all attacks with an image quality fixed at 30dB. The attacks reduce the detection rate considerably, demonstrating that deep fakes can be manipulated with fingerprint information only, so that they are classified as actual images. 

\paragraph{Attack Analysis}
To gain more insights into these results, we first examine the \emph{frequency-bars attack} and its effectiveness against the considered detection methods.
Despite its simplicity, the attack is highly successful against the CNN-based classifier and the regression model by \citeauthor{frank_leveraging_2020} These results suggest that the two classifiers mainly rely on information stored in the high-frequency bands for their decisions. In contrast, the attack only provides low success rates against the detector of \citeauthor{joslin_attributing_2020}, indicating that low-frequency artifacts are also relevant in the approach.

Interestingly, we obtain the exact opposite results for the \emph{mean-spectrum attack}.
This attack works considerably well against \citeauthor{joslin_attributing_2020} and can precisely remove the detected pattern.
However, it fails to circumvent the classifiers by \citeauthor{frank_leveraging_2020}
We attribute the low success rate to the fact that the classifiers operate on log-scaled spectra, while the attack only performs non-scaled manipulations, thus ignoring the peculiarities of the classifiers.

This intuition is further strengthened by the results obtained for the \emph{peak-extraction attack}, which relies on the log-scaled spectra to calculate the fingerprints.
The success of the peak-extraction attack, however, depends on the setup: it works almost perfectly against ProGAN and SNGAN, which show strong peaks throughout the spectrum.
To confirm that the extracted peaks are accurate for each GAN instance, we perform an additional experiment, in which we cross-remove the fingerprint of individual GAN-instances from images of other GANs.
Indeed, we find that removing their own fingerprint results in a more successful attack for each classifier. 

To our own surprise, the \emph{regression-weights attack} is rarely successful---even against a regression model.
Our analysis shows that the computed fingerprints exhibit patterns across the entire frequency spectrum,
so that the attack also manipulates lower frequency bands. While effective as attacks alone, these manipulations lead to a substantial decrease in image quality and weaken the overall performance.

\subsection{Image Quality}
\label{subsec:eval-image-quality}

The manipulations performed by our counterattacks in the frequency domain may lead to visible artifacts in the spatial domain. As these artifacts might reveal the attack and provide new ground for detection, we also analyze how much the different counterattacks affect the overall image quality.

\begin{figure}[] \subfloat[Original]
	{\includegraphics[width=.18\linewidth]{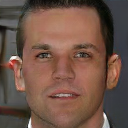}}\hfill
	 \subfloat[Mean] 
	{\includegraphics[width=.18\linewidth]{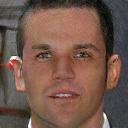}}\hfill
	 \subfloat[Peaks] 
	{\includegraphics[width=.18\linewidth]{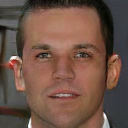}}\hfill
	 \subfloat[Regress.] 
	{\includegraphics[width=.18\linewidth]{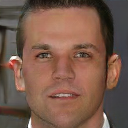}}\hfill
	 \subfloat[Bars] 
	{\includegraphics[width=.18\linewidth]{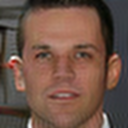}}
	\\
	\subfloat[Original]
	{\includegraphics[width=.18\linewidth]{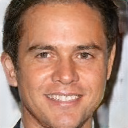}}\hfill
	 \subfloat[Mean] 
	{\includegraphics[width=.18\linewidth]{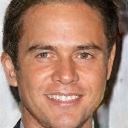}}\hfill
	 \subfloat[Peaks] 
	{\includegraphics[width=.18\linewidth]{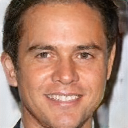}}\hfill
	 \subfloat[Regress.] 
	{\includegraphics[width=.18\linewidth]{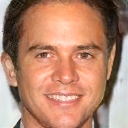}}\hfill
	 \subfloat[Bars] 
	{\includegraphics[width=.18\linewidth]{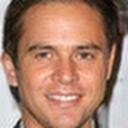}}\hfill	
	 \caption{Modified deep-fake examples from CelebA SNGAN (a-e) and CelebA ProGAN (f-j). All attacks 
	are performed with a fixed image quality of 30dB.} 
	\label{fig:evaluation-selection-examples-2} 
	\end{figure}

\autoref{fig:evaluation-selection-examples-2} shows two representative examples of deep-fake images modified by the different counterattacks at a fixed \ac{psnr} of roughly 30dB.
While all attacks affect the image quality only slightly, the peak extraction preserves the image details particularly well.
Note, however, that this attack yields only moderate success rates (see Table~\ref{tab:attack-success}).
The frequency-bars attack, in contrast, introduces more visible artifacts, but the attack also provides good results despite its simplicity, achieving the highest success rates against two of the detection methods.
Moreover, we find that its high success rates remain stable even for better \ac{psnr} values of up to 37dB, where artifacts are rarely visible anymore.

Overall, these results show that all attack variants are effective with minor impact on the visual quality in most cases. 
Even for the frequency-bars attack, its impact on the image quality is acceptable on the examined data.

\subsection{Comparison to Image Perturbations}
\label{subsec:evaluation-image-perturbations}
In the next experiment, we compare our counterattacks with image perturbations that prior work used to test the robustness of detection methods~\cite{frank_leveraging_2020, yu_attributing_2019, joslin_attributing_2020}. In particular, we implement cropping, noise addition, blurring, and JPEG compression.  
We again execute the perturbations with such a strength that the image quality drops to about 30dB on average.

\autoref{tab:attack-success} shows attack success rates of the considered image perturbations. 
While blurring also achieves a high success rate across all settings, the other perturbations show mixed success rates that depend on the respective dataset, detection method, and GAN class/model. In comparison, our counterattacks are more effective, which motivates their usage as additional baselines in future work. 

\subsection{Summary of Results}

In summary, our experiments demonstrate the effectiveness of the proposed counterattacks. The different variants outperform previous perturbation attacks, without affecting the visual quality significantly at the same time. However, we find that their success depends on various factors, so that a single universal attack strategy does not exist. For instance, while the \emph{mean-spectrum attack} yields the highest success rate against the detection method by \citet{joslin_attributing_2020}, it is largely ineffective against the other two detectors, where the \emph{frequency-bars attack} and \emph{peak-extraction attack} are most effective here.

\section{Discussion}
\label{sec:discussion}
Our evaluation demonstrates how our simple counterattacks impact various deep-fake detectors. Still, there are open questions that we discuss in the following. 

\begin{figure}[]
	\settoheight{\tempdima}{\includegraphics[width=.22\linewidth]{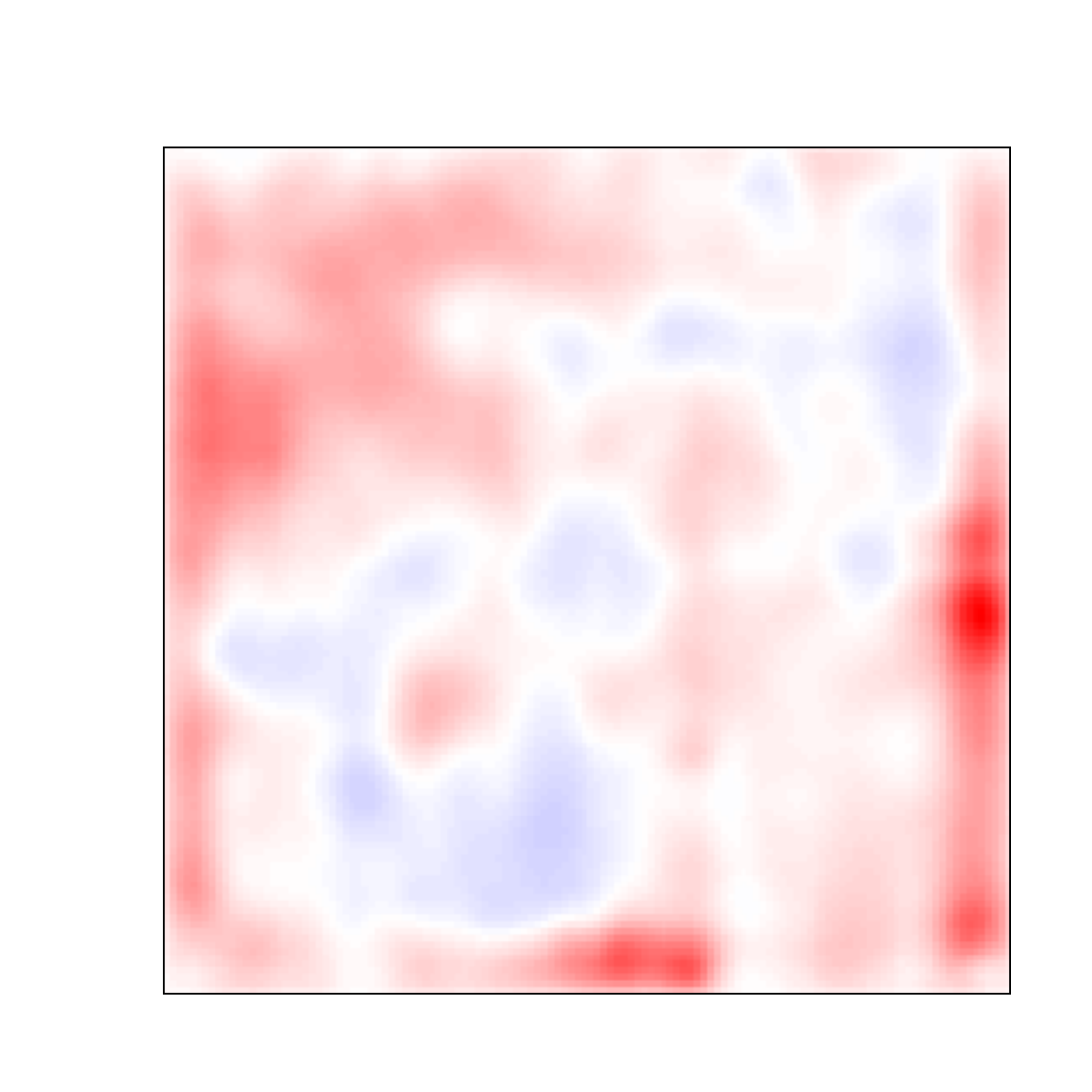}}%
	\centering
	\begin{tabular}{@{}c@{ }c@{ }c@{ }c@{ }c@{}}
		&\textit{ProGAN} & \textit{SNGAN} & \textit{CramerGAN} & 
		\textit{MMDGAN}\\
		\rowname{LSUN}&
		\includegraphics[width=.22\linewidth]{./images/explanations/progan/lsun_lrp__sequential_preset_a_flat_weights_frank_correct_converted.pdf}&
				\includegraphics[width=.22\linewidth]{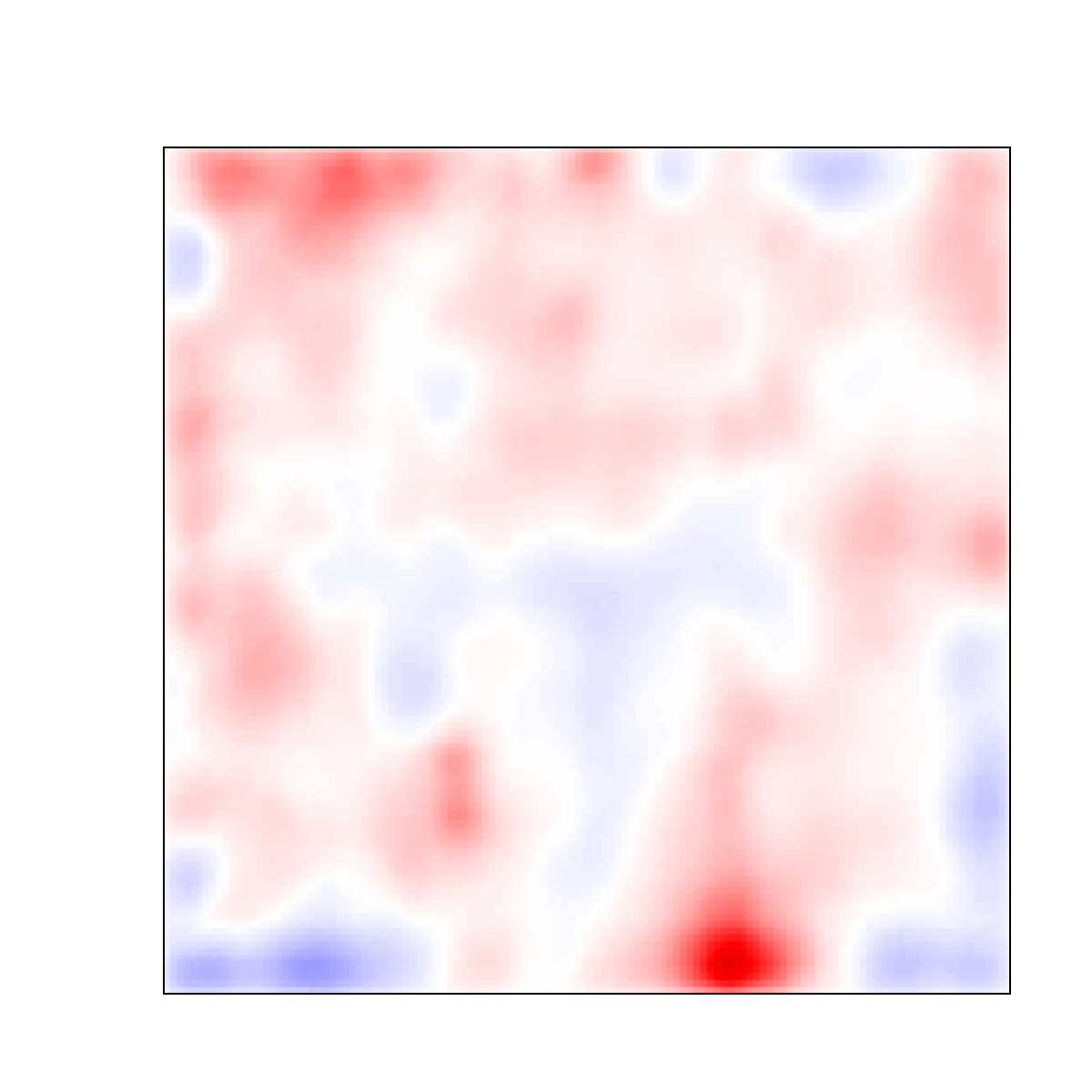}&
		\includegraphics[width=.22\linewidth]{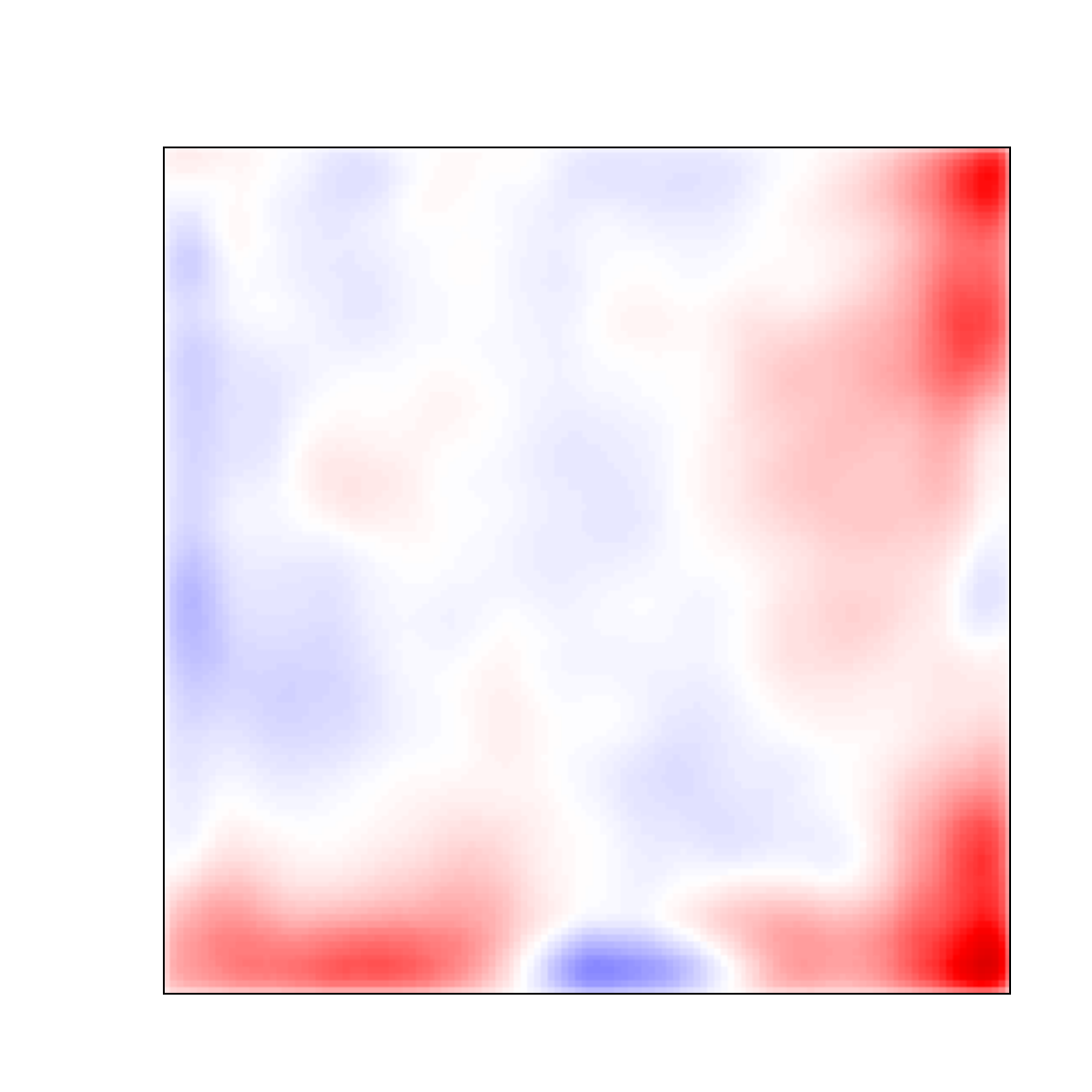}&
		\includegraphics[width=.22\linewidth]{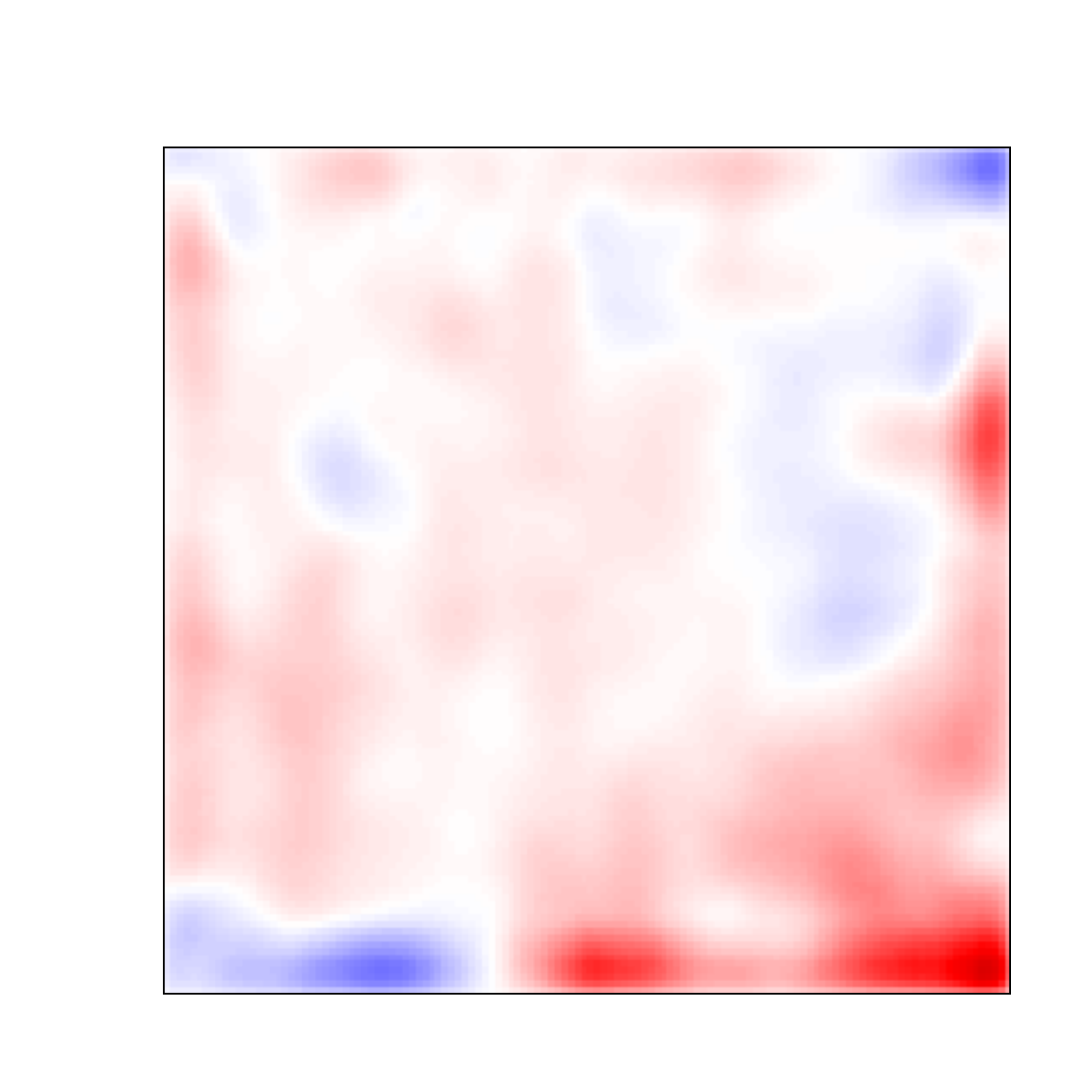}\\[-1ex]
		\rowname{CelebA}&
		\includegraphics[width=.22\linewidth]{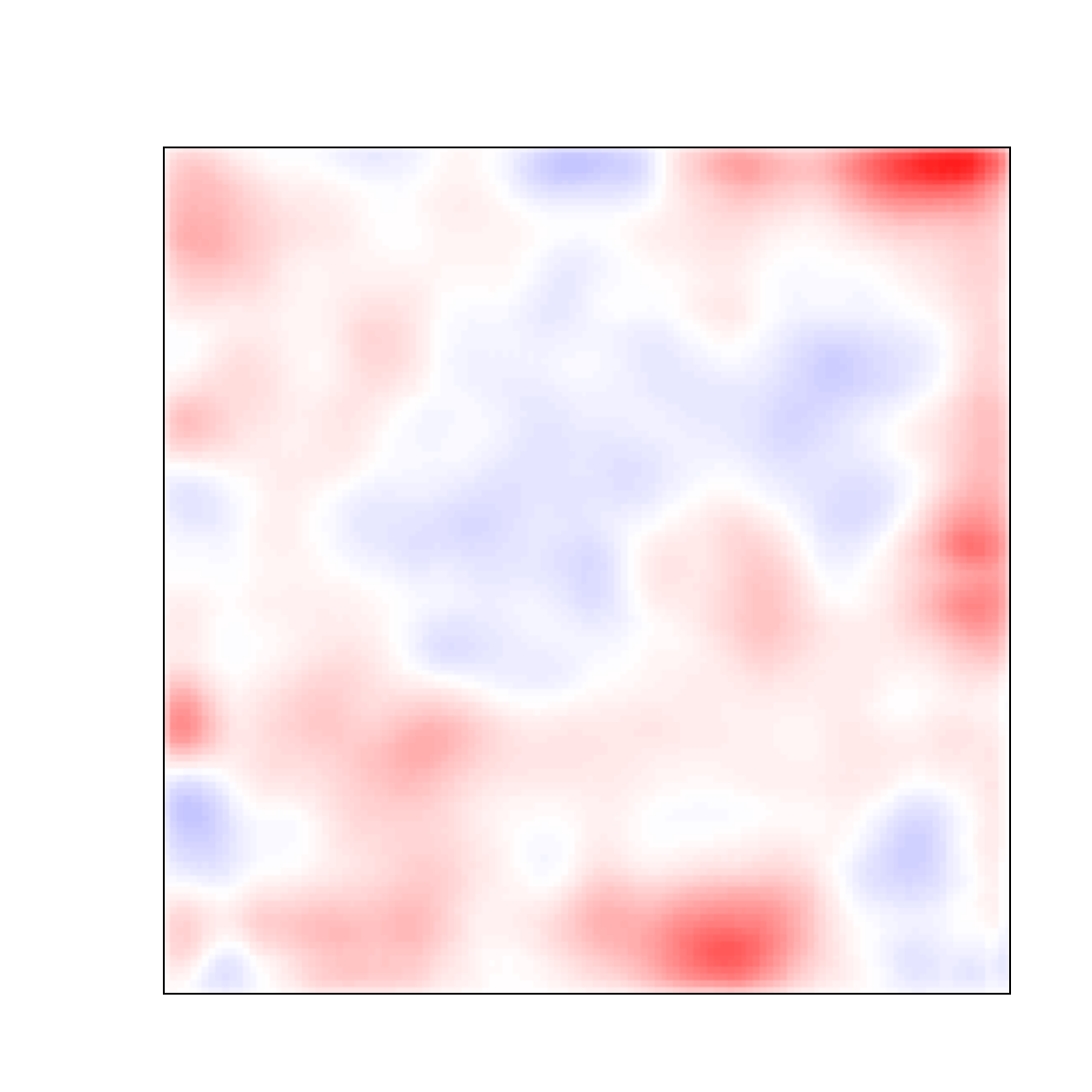}&
		\includegraphics[width=.22\linewidth]{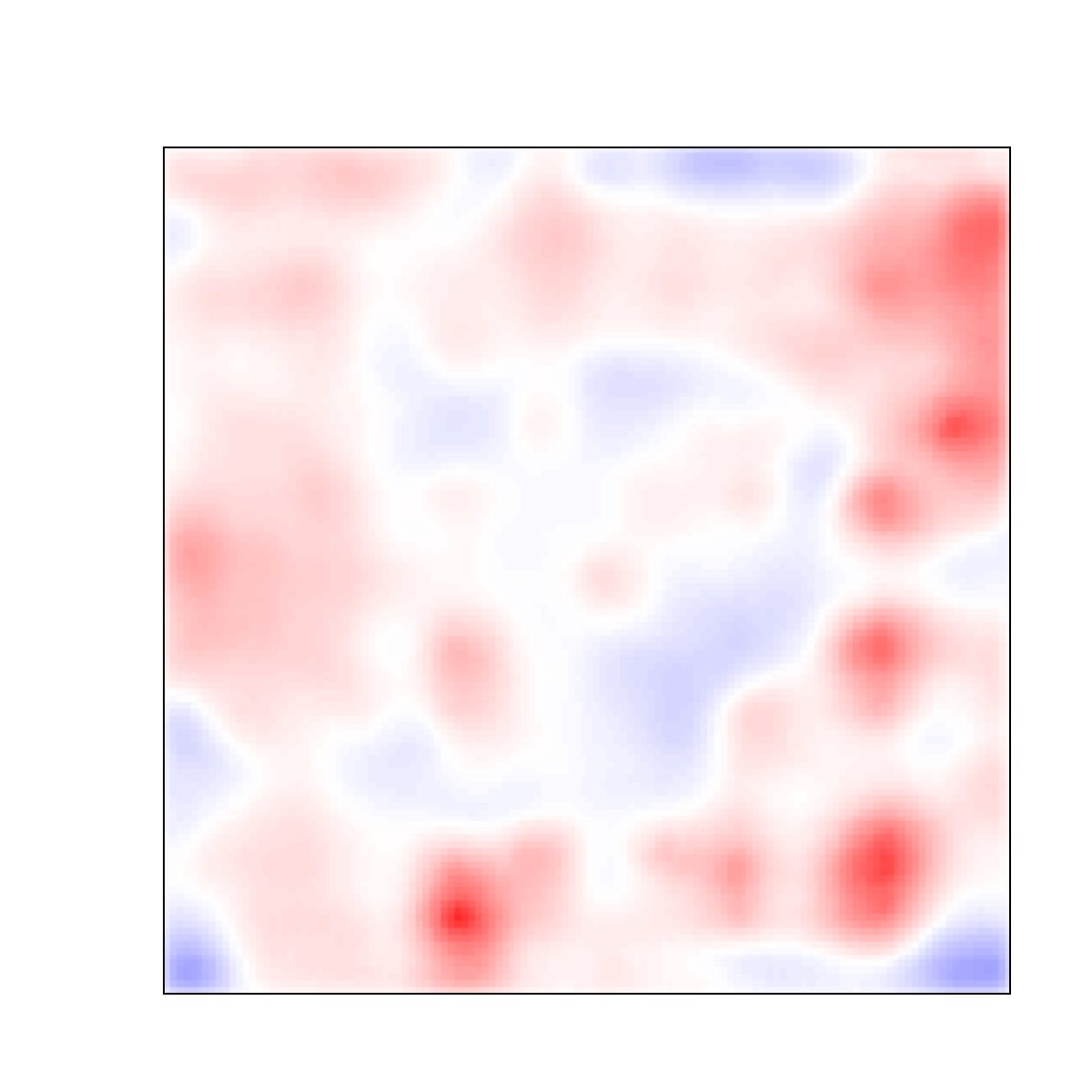}&
		\includegraphics[width=.22\linewidth]{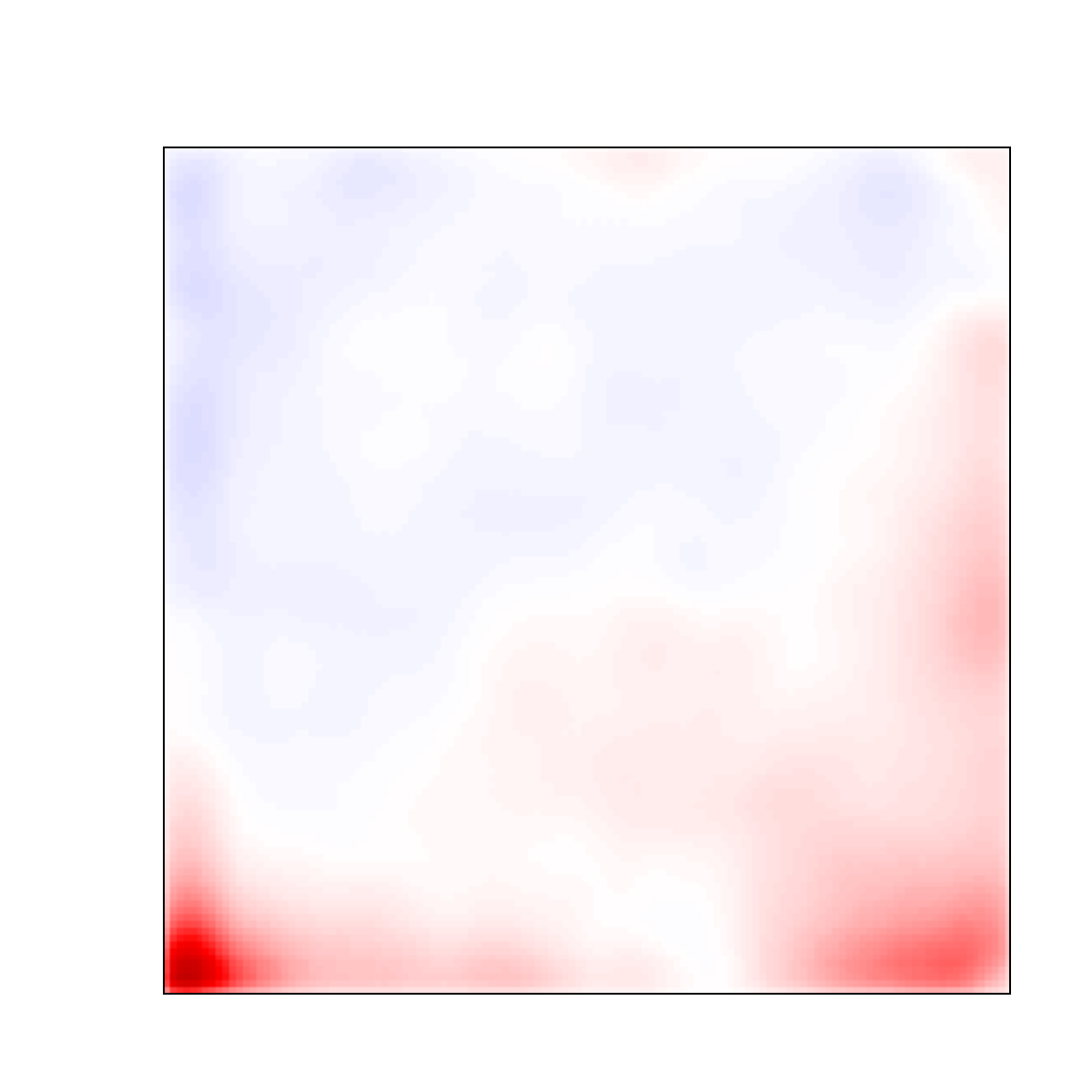}&
		\includegraphics[width=.22\linewidth]{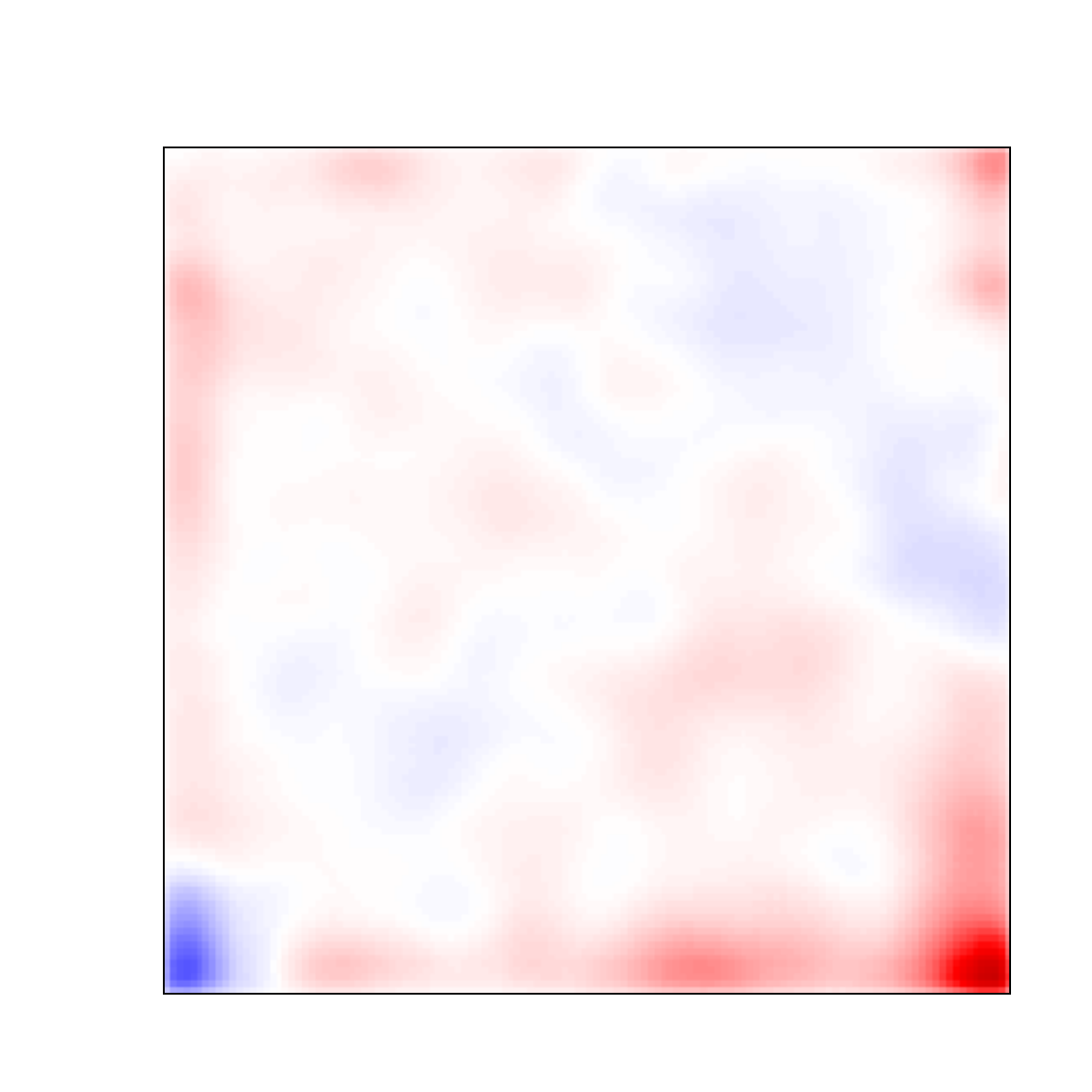}\\[-1ex]
	\end{tabular}	
	\caption{Mean LRP-explanations in the frequency spectrum for the CNN detector~\cite{frank_leveraging_2020}. Red areas correspond to a positive, blue areas to a negative contribution to the deep-fake prediction.
	} %
	\label{fig:discussion-explanations}
\end{figure}

\paragraph{A Closer Look on the Frequency Spectrum}
Although our presented attacks allow evading the detection in most cases, there is no universal method that is successful in any setup. 
The success rate can even vary between the different GAN architectures for a respective combination of dataset, detection method, and attack.
To better understand the results, we therefore explain the predictions using the example of the CNN-based classifier~\cite{frank_leveraging_2020}. We apply LRP, which is a well-established method for analyzing the decisions of various deep neural network architectures~\cite{BacBinMon+15}.
\autoref{fig:discussion-explanations} depicts the explanations averaged over 1,000 unmodified deep-fake images. 

Our analysis shows that the explanations for the different GAN models and datasets vary---supporting the concept of characteristic GAN artifacts~\cite{frank_leveraging_2020}.
Amongst others, we find that the relevance of specific frequency bands differs between individual GAN models: while, for instance, the classifier seems to consider the whole spectrum in the case of ProGAN and SNGAN, it mainly focuses on higher frequencies for CramerGAN and MMDGAN. 
Similarly, the focus on the frequency bands even appears to vary between the datasets for a particular GAN.
This finding might also explain the differences we experience between the results on these datasets, as even the same counterattack might yield different success rates depending on the given data (see \autoref{sec:evaluation}).

\paragraph{Limitations}
We leave counter-defenses to our attacks, the next step in the arms race of attackers and defenders, to future work. For instance, our modified deep-fake images could be added to the training process of deep-fake detectors, similar to adversarial training~\cite{GooShlSze15}. Ultimately, iterative research moving back and forth between attacks and defenses likely enables deeper insights on the characteristics of GAN models.

Moreover, we solely focus on frequency-based detectors, which outperform approaches in the spatial domain~\cite{frank_leveraging_2020}.
However, our preliminary results on attacks against the approach by \citet{yu_attributing_2019} indicate that frequency-based attacks might be less effective against spatial detection methods.
This insight motivates research on fingerprint-based counterattacks in the spatial domain, which we also leave to future research.

\section{Related Work}
The evasion of deep-fake detection methods is an active area of research that can be divided into the following strains: 
First, an adversary can create an adversarial example of the deep-fake image of interest~\cite{carlini_evading_2020, liaoImperceptibleAdversarialExamples2021, neekharaAdversarialThreatsDeepFake2020}.
Second, the training of the GAN can be directly adapted~\cite{durall_watch_2020, jungSpectralDistributionAware2020}.
For example, \citet{durall_watch_2020} show that common upsampling methods prevent models from reproducing the spectral distribution of natural images in the GAN images. Thus, they introduce a spectral regularization term that trains spectrally consistent GANs. 

Another line of attacks uses learning-based systems to modify deep fakes~\cite{cozzolino_spoc_2019, wangAdversarialAttackFakeFaces2021, zhaoMakingGANGeneratedImages2021}. For example, \citet{cozzolino_spoc_2019} train a GAN to insert the fingerprint of a camera into GAN-generated images while removing the own GAN fingerprint. 
\citet{neves_ganprintr_2020} target high frequencies by using an auto-encoder that encodes an image into a smaller dimensional space before decoding it again, thereby removing unimportant information.

However, \citet{huang_fakepolisher_2020} state that methods, such as \citet{cozzolino_spoc_2019} and \citet{neves_ganprintr_2020}, introduce new artifacts when removing fingerprints. Hence, they propose a shallow reconstruction by learning a dictionary model on natural images, which is a low-dimensional subspace representing these images. A deep-fake is mapped to a representation in the subspace and then reconstructed. 

Our approach represents a novel class of attacks. 
We directly manipulate the frequency spectrum of deep fakes by targeting a GAN fingerprint.
The attack operates in a black-box scenario with access to GAN images only. In contrast to prior work, our attacks are conceptually simple and do not require adjusting GANs or training sophisticated learning-based systems.

\section{Conclusion}
This paper presents a novel class of simple attacks for bypassing deep-fake detection. The attacks remove GAN artifacts from images directly in the frequency spectrum. 
Our evaluation shows that depending on the combination of dataset, GAN, and detection method, an adversary can use one of our attacks to mislead the detection. 
In conclusion, we thus provide evidence that current approaches for detecting deep-fake images are still far from robust and can be evaded easily.

\section*{Acknowledgments}
We would like to thank Thorsten Holz and Joel Frank for the valuable insights and discussions regarding the presented research idea.
The authors gratefully acknowledge funding from the German Federal Ministry of Education and Research (BMBF) under the project BIFOLD (Berlin Institute for the Foundations of Learning and Data, ref.  01IS18025A and ref 01IS18037A) 
and from the Deutsche Forschungsgemeinschaft (DFG, German Research Foundation) 
under Germany's Excellence Strategy EXC 2092 CASA-390781972 and the projects 456292433; 456292463.

{\footnotesize \balance
\bibliographystyle{abbrvnat} %

}

\end{document}